\documentclass{article}

\PassOptionsToPackage{numbers}{natbib}

\usepackage[preprint]{neurips_2024}

\usepackage[utf8]{inputenc} 
\usepackage[T1]{fontenc}    
\usepackage{hyperref}       
\usepackage{url}            
\usepackage{booktabs}       
\usepackage{amsfonts}       
\usepackage{nicefrac}       
\usepackage{microtype}      
\usepackage{xcolor}         

\usepackage{bbm} 
\usepackage{booktabs} 
\usepackage{subcaption} 
\usepackage{graphicx} 
\usepackage{enumitem} 
\usepackage{float} 

\title{WV-Net: A foundation model for SAR WV-mode satellite imagery trained using contrastive self-supervised learning on 10 million images}

\author{
  Yannik Glaser\\
  Information and Computer Sciences\\
  University of Hawai`i at M\=anoa\\
  Honolulu, HI 96822 \\
  \texttt{yglaser@hawaii.edu} \\
  \And
  Justin E. Stopa\\
  Ocean Resources and Engineering\\
  University of Hawai`i at M\=anoa\\
  Honolulu, HI 96822 \\
  \texttt{stopa@hawaii.edu} \\
  \And
  Linnea M. Wolniewicz\\
  Information and Computer Sciences\\
  University of Hawai`i at M\=anoa\\
  Honolulu, HI 96822 \\
  \texttt{linneamw@hawaii.edu} \\
  \And
  Ralph Foster\\
  Applied Physics Laboratory\\
  University of Washington\\
  Seattle, WA 98105 \\
  \texttt{rcfoster@uw.edu} \\
  \And
  Doug Vandemark\\
  Ocean Processes Analysis Laboratory\\
  University of New Hampshire\\
  Durham, NH 03824 \\
  \texttt{doug.vandemark@unh.edu} \\
  \And
  Alexis Mouche\\
  Univ. Brest, CNRS, IRD, Ifremer, Lab of Ocn. Phys.\\
  IUEM\\
  29280, Brest, France \\
  \texttt{Alexis.Mouche@ifremer.fr} \\
  \And
  Bertrand Chapron\\
  Univ. Brest, CNRS, IRD, Ifremer, Lab of Ocn. Phys.\\
  IUEM\\
  29280, Brest, France \\
  \texttt{Bertrand.Chapron@ifremer.fr} \\
  \And
  Peter Sadowski\\
  Information and Computer Sciences\\
  University of Hawai`i at M\=anoa\\
  Honolulu, HI 96822 \\
  \texttt{peter.sadowski@hawaii.edu} \\
}

\begin{document}

\maketitle

\vspace{-0.1in}  
\raggedbottom  
\begin{abstract}
    The European Space Agency’s Copernicus Sentinel-1 (S-1) mission is a constellation of C-band synthetic aperture radar (SAR) satellites that provide unprecedented monitoring of the world’s oceans. S-1's wave mode (WV) captures 20x20 km image patches at 5 m pixel resolution and is unaffected by cloud cover or time-of-day. The mission’s open data policy has made SAR data easily accessible for a range of applications, but the need for manual image annotations is a bottleneck that hinders the use of machine learning methods. This study uses nearly 10 million WV-mode images and contrastive self-supervised learning to train a semantic embedding model called WV-Net. In multiple downstream tasks, WV-Net outperforms a comparable model that was pre-trained on natural images (ImageNet) with supervised learning. Experiments show improvements for estimating wave height (0.50 vs 0.60 RMSE using linear probing), estimating near-surface air temperature (0.90 vs 0.97 RMSE), and performing multilabel-classification of geophysical and atmospheric phenomena (0.96 vs 0.95 micro-averaged AUROC).  WV-Net embeddings are also superior in an unsupervised image-retrieval task and scale better in data-sparse settings. Together, these results demonstrate that WV-Net embeddings can support geophysical research by providing a convenient foundation model for a variety of data analysis and exploration tasks. 
\end{abstract}

\section{Introduction}

Machine learning is becoming increasingly important for analyzing remote sensing data. The number of Earth observation satellites in orbit has grown from 150 in 2008~ \cite{eosats_tatem_2008} to over 1150 in 2022~\cite{ucs_sat_db_2022}. Missions like the European Space Agency's (ESA) Sentinel-1 (S-1) mission generate large amounts of high-resolution images with global coverage. ESA has taken an open-data policy making high-resolution synthetic aperture radar (SAR) imagery readily available for applications ranging from environmental monitoring to climate modeling~\cite{TorresEtAl2012}. Fully leveraging the torrent of S-1 SAR imagery requires automated analysis tools with many potential applications for machine learning~\cite{TopouzelisKitsiou2015,WangEtAl2019}. However, the machine learning approach generally requires large datasets of training images that have been annotated by experts.

Transfer learning is a common solution to this challenge. A deep neural network model is first pretrained on a large dataset from a related domain and then fine-tuned on the target task, requiring significantly less labeled data than would be necessary when training from randomly initialized network parameters. The pretrained model is called a \textit{foundation model} because it can be reused for multiple downstream tasks. Foundation models pretrained to classify natural images (primarily the ImageNet dataset~\cite{imagenet_deng_2009}) are routinely fine-tuned for remote tasks~\cite{transfersceneclsrs_hu_2015, wvmodelorig_wang_2018, transfersceneclsrs_nogueira_2015}. However, transferring a from a natural image classification task to a remote sensing task can be problematic because the image characteristics are so different. This is known as the domain gap and the deep learning literature has repeatedly shown that a wide domain gap between pretraining and target data domains can hinder transfer performance~\cite{transferability_yosinski_2014, quantmedss_leong_2022, imnettransfer_raghu_2019}. 

Self-supervised learning (SSL) provides an alternative approach to pretraining a foundation model with unannotated, domain-specific data. Instead of predicting annotations in a supervised manner, SSL algorithms define some other \textit{pretext task} for pretraining. This approach has long been utilized in natural language processing~\cite{word2vec_church_2017, glove_pennington_2014} and has been one of the driving factors for the success of large language models~\cite{bert_devlin_2018, nlpgenpretraining_radford_2018} resulting in tools like ChatGPT~\cite{gpt3_brown_2020}. Recently, contrastive learning has re-emerged as a successful self-supervised form of pretraining, especially for computer vision~\cite{simclr_chen_2020, dino_caron_2021, ibot_zhou_2022}. Contrastive algorithms have produced impressive results on natural image datasets, resulting in general-purpose network weights that perform on par or better than supervised networks being trained from scratch on the target dataset\cite{dino_caron_2021, simclr_chen_2020}. Thus, SSL presents opportunities for analyzing remote sensing data. Recent studies have shown that pretraining on remote sensing data instead of natural images yields superior performance on downstream tasks~\cite{rsdomainpretraining_neumann_2019}, with most proposed methods being self-supervised~\cite{satvit_fuller_2022, satmae_cong_2022, scalemae_reed_2023, geoawaressl_ayush_2021}. To date, these efforts have focused on remote sensing imagery of landmasses or coastal regions. 

The objective of this work is build the first foundation model for open-ocean sea surface images. Our foundation model is pretrained on imagery from S-1 WaVe (WV) mode, which was designed to capture ocean waves at 5 m resolution in 20x20 km footprints~\cite{HasselmannEtAl2013,cwaves1_stopa_2017, waveheight_quach_2020}. These images capture a variety of ocean phenomena~\cite{JacksonApel2004} and have global coverage, with millions of images archived over the last decade. Thus, the data has been used to study ocean fronts~\cite{RascleEtAl2017}, air-sea interactions including organized turbulence in the marine boundary layer~\cite{YoungEtAl2000,VandemarkEtAl2001,StopaEtAl2022}, and other physical, atmospheric, and biological processes~\cite{WangEtAl2019a}. By building a foundation model specific to SAR WV images, we hope to accelerate this research.

Two hypotheses are tested. First, we test whether contrastive SSL can train a SAR WV-mode foundation model that outperforms standard computer vision models pretrained on natural images. Second, we test whether performance of the model can be improved by using domain knowledge to design data augmentations. These hypotheses are tested experimentally using a dataset of almost 10 million S-1 WV images, along with three smaller subsets of annotated images that exemplify target supervised tasks for transfer learning. The optimized foundation model is made publicly available under the name WV-Net, and we expect it to be useful for a variety of downstream applications such as studying air-sea interactions, improving constraints on numerical weather predictions, and monitoring sea ice.

\section{Methods}

\subsection{Datasets}

The S-1 mission has been operating two SAR satellites for almost a decade. The satellites are equipped with C-band instruments that operate continuously day and night, unaffected by cloud cover. While S-1 has four imaging modes, we focus exclusively on the WV mode that is used over open ocean. Each satellite produces approximately 60,000 WV images per month, and in total there are approximately 165 months and 9.9M images. Images are stored as \textsc{png} files with 20x20 km footprints and 5 m resolution. They contain features of interest at multiple spatial scales (Figure~\ref{fig:sar_augmentations}), sometimes in the same image. Image preprocessing is detailed in Appendix~\ref{appendix_sar}. Below we describe three subsets of the data that have been annotated for supervised learning tasks; these tasks are used to evaluate WV-Net as a foundation model.

\paragraph{GOALI classification dataset}
The GOALI dataset~\cite{goali_unpublished} consists of 10,000 WV images that were manually annotated by human experts in SAR imagery for multilabel classification (an image can have multiple labels at once). The labels indicate geophysical phenomena observable in the image. To this we add 6,400 images from~\citet{WangEtAl2019a} that have been re-annotated in a way consistent with GOALI, for a total of annotated 16,400 images. The GOALI images are multilabeled with the following phenomena: wind streaks (WS), micro-scale convective cells (MC), negligible atmospheric variability (NV), rain cells (RC), cold pools (CP), sub-mesoscale air-mass boundaries (AB), low wind areas (LW), atmospheric gravity waves (AW), biological slicks (BS), ocean fronts (OF), internal oceanic waves (IW), icebergs (IB), ships (SH), ship wakes (SW), and other unidentified phenomena (UD). Sample images are shown in Figure~\ref{fig:sar_augmentations} while a representative example from each class is shown in Figure~\ref{fig:sar_examples} of the appendix. This dataset is currently unpublished work but will be made publicly available in the future. 

\begin{figure}
\centering
\includegraphics[width=33pc]{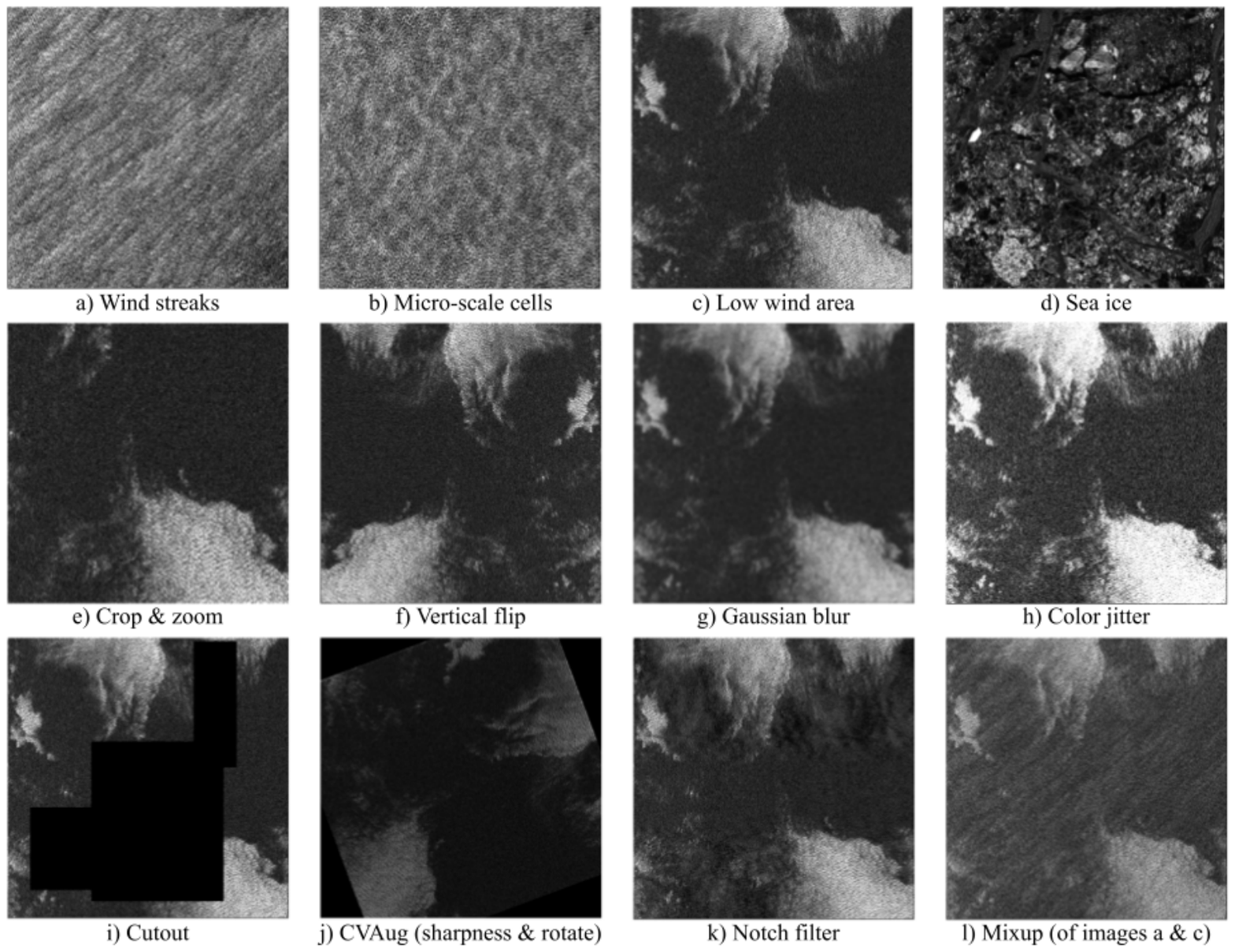}
\caption{(a--d): Sample images of different geophysical phenomena observable in the global S-1 WV archive, titled by their dominant classes. Multiple classes can be present in the same image. (e--h): Augmented versions of the low wind image illustrating the default SimCLR augmentation policies. (i--l): Augmented low wind images illustrating the augmentation policies evaluated in this work. In the actual SimCLR framework, usually multiple augmentation are applied in sequence to the same image.}
\label{fig:sar_augmentations}
\end{figure}

\paragraph{Wave height regression dataset}  
\citet{waveheight_quach_2020} annotated hundreds of thousands of WV images with significant wave height ($H_{s}$) by colocating S-1 satellites with altimeter satellites. Here we use a subset of 200,000 images and randomly split the data into sets of 50,000, 50,000, and 100,000 for training, validation, and final evaluation, respectively.

\paragraph{Air temperature regression dataset}
\citet{StopaEtAl2022} showed that the sea surface roughness observed in SAR is related to atmospheric stratification and therefore air-sea temperature differences. 
Using ERA5 reanalysis data~\cite{HersbachEtAl2020} as ground truth annotations for the sea surface temperature (SST) and air temperature ($T_{v10}$), we attempt to predict the difference from SAR images. 
The air temperature is corrected using the COARE algorithm~\citet{EdsonEtAl2013} to account for moisture content and called a virtual air temperature. The annotated dataset consists of 76,000 images, which is split into 50,000 training, 11,000 validation, and 15,000 testing images.

\subsection{Implementation details}

\begin{figure}[ht!]
  \centering
  \includegraphics[width=0.6\linewidth]{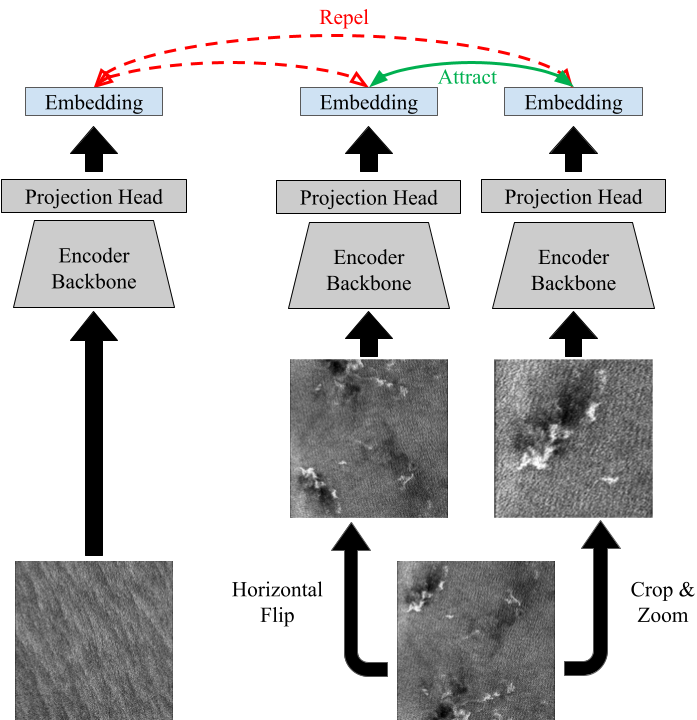}
  \caption{In the SimCLR algorithm, images are randomly augmented to create several views of the same image. An encoder network --- consisting of a backbone and a smaller projection head --- learns to produce an embedding that is similar to embedded views from the same original image and dissimilar to embedded views from all other images. Only the encoder backbone is used for transfer learning.}
  \label{fig:Fig02}
\end{figure}

WV-Net is trained using the SimCLR contrastive SSL framework~\cite{simclr_chen_2020} with a ResNet50 backend architecture~\cite{resnet_he_2015}. These choices were based on an initial exploratory analysis of framework and backend combinations detailed in Appendix~\ref{appendix_contrastive}. The SimCLR SSL pretext task is to learn similar representations for two augmented \textit{views} of an image while discouraging similarity with the representations of any other image in the training data (Figure~\ref{fig:Fig02}). 

A SimCLR training step begins by randomly sampling a mini-batch of $N$ training images. Each image $\mathbf{x}_k$ is transformed twice by random sequences of augmentation policies (sampled from a pool of transformations) to produce two views of the original image, $\mathbf{\tilde{x}}_{2k-1}$ and $\mathbf{\tilde{x}}_{2k}$, resulting in 2N total images. Each view is encoded by a backend network (here a ResNet50) and then a smaller \textit{projector} neural network, resulting in the embedding vectors $\mathbf{z}_{2k-1}$ and $\mathbf{z}_{2k}$. The loss for any \textit{positive pair} of embeddings $\mathbf{z}_{i}$ and $\mathbf{z}_{j}$ originating from the same image is:
\begin{equation}
    l_{i,j} = -\log \frac{\exp(sim(\mathbf{z}_i, \mathbf{z}_j)/\tau)}{\sum^{2N}_{k=1}\mathbbm{1}_{[k\neq i]}\exp(sim(\mathbf{z}_i, \mathbf{z}_k)/\tau)}
\end{equation}
where $\tau$ is a temperature scalar, $\mathbbm{1}$ is the indicator function and  $sim(\cdot,\cdot)$ is the cosine similarity:
\begin{equation}
    sim(\mathbf{u},\mathbf{v}) =\frac{\mathbf{u}^{T}\mathbf{v}}{\Vert\mathbf{u}\Vert \Vert\mathbf{v}\Vert}
    \label{cos_sim_eq}
\end{equation}
Unless otherwise specified, all hyperparameters are adopted from the original SimCLR work~\cite{simclr_chen_2020} with linear learning rate scaling. All deep learning models are implemented using a combination of PyTorch~\cite{pytorch_paszke_2019} and PyTorch Lightning~\cite{pytorchlightning_falcon_2019}, while other machine learning models used for transfer learning are implemented in scikit-learn~\cite{sklearn_pedregosa_2011}.

\subsection{Augmentations}

SAR WV mode images are very different from natural images, so experiments were conducted to optimize the choice of augmentations used to train WV-Net. In addition to the augmentations proposed in the original SimCLR~\cite{simclr_chen_2020}, we explore a variety of augmentations proposed in the contrastive learning literature, transformations from traditional computer vision, and a transform from signal processing that was inspired by the SAR imaging process. 

\begin{itemize}
\item \textbf{SimCLR augmentations:} These include random cropping and zooming, random flipping, random color jitter, and random Gaussian blur (see Figure~\ref{fig:sar_augmentations}e-h for examples).  
\item \textbf{Literature-inspired augmentations:} \textit{Mixup}~\cite{mixup_zhang_2018} and \textit{Cutout}~\cite{cutout_devries_2017} are policies that have been shown to work well in contrastive learning frameworks~\cite{plasticcutout_jin_2024, simclr_chen_2020, dacl_verma_2021}.
\item \textbf{Computer vision augmentations:} Many traditional image processing transformations seem well-suited for this application. We combined random rotation, random color inversion, and random sharpness transformations into a single augmentation policy called \textit{CVAug}. We also modified the crop-and-zoom augmentation that is universal among multi-view contrastive learning frameworks to create a \textit{no-zoom crop} policy that focuses on random cropping with only minimal scaling. Since WV images are captured from a satellite in constant orbit and have a consistent 20km footprint, phenomena captured don't vary in scale as much as features in natural images might. Thus, by reducing the zoom component, scale invariance is not as heavily incentivized in the model allowing features to be more specific.
\item \textbf{Domain-inspired augmentation:} WV images are often dominated by ocean surface waves, so representing the image in the frequency domain and dropping random frequency components emphasizes or de-emphasizes particular features that could be relevant to sea-surface state. This is a common signal-processing operation called \textit{random notch filtering}.
\end{itemize}

Examples for each augmentation policy can be seen in Figure~\ref{fig:sar_augmentations} and details for the augmentation policy implementations and more rationales on each policy's inclusion are provided in the Appendix~\ref{appendix_augmentations}.
All augmentations are added to the overall transform pool from which to sample during training. That means that each augmentation policy, be it from the original SimCLR policies or one of the added policies described here, gets applied with some probability to each image. An image may be transformed by any combination of policies, including all or none, and the sampling is repeated for every image in every batch. 

\subsection{Evaluation protocols}

To evaluate the quality of the WV-Net embeddings, we conduct experiments in which the embeddings are used for a multilabel classification task, two regression tasks, and an unsupervised image retrieval task. The experiment protocols are summarize below with more details in Appendix~\ref{appendix_eval}.

\paragraph{Multilabel classification}
Four common protocols are used to evaluate the embeddings for transfer learning to a multilabel classification task: the $k$-nearest neighbors (kNN) approach from~\citet{instancedisc_wu_2018}, the linear probing protocol from~\citet{simclr_chen_2020}, the multilayer-perceptron (MLP) probing protocol suggested by~\citet{democratizingjessl_bordes_2023}, and full end-to-end finetuning following recent trends in evaluation~\cite{mae_he_2021, beit_bao_2022, ibot_zhou_2022, prec_dong_2022}. The primary metric used for evaluation is the micro-averaged area under the receiver operating characteristic curve (AUROC). The F1-score is also reported.

\paragraph{Regression}
Three protocols are used for two regression tasks: linear probing, MLP probing, and end-to-end finetuning. These protocols are identical to the classification protocols except no kNN-based model is considered. Models are evaluated using the mean absolute error (MAE) and root mean squared error (RMSE). 

\paragraph{Image retrieval}
The embeddings are evaluated for one-shot image retrieval, following the kNN-retrieval approach from~\citet{dino_caron_2021}. Experiments are conducted on the rarest classes from the combined classification dataset, occurring in no more than 1,000 images ($<$0.05\% of the total dataset), consisting of seven total classes. Models are evaluated in terms of Mean average precision (mAP) averaged over all classes.


\section{Results}

Experiments were conducted to test two hypotheses: (1) a self-supervised model trained on WV data will outperform a model trained on ImageNet, and (2) the self-supervised model can be improved by selecting pretext tasks based on domain-specific properties of the satellite images.
We first performed experiments to optimize the set of augmentations used in the SimCLR training algorithm. WV-Net was then trained using the optimized augmentations for an extended period. This model is compared to an ImageNet model and a WV model trained with the default SimCLR augmentations, testing hypotheses (1) and (2), respectively.

\subsection{Optimization of WV-mode specific data augmentations}

\begin{figure}[ht!]
    \centering
    \includegraphics[width=0.9\linewidth]{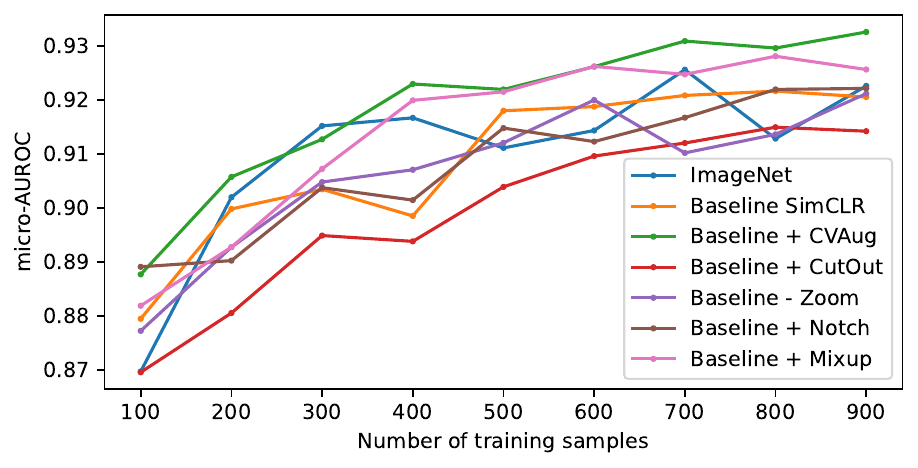}
    \caption{Performance of various embeddings (Micro-AUROC, higher is better) vs. number of labeled training samples in the multilabel classification task. This experiment used the MLP transfer learning protocol.}
    \label{fig:auroc_samples_classification}
\end{figure}

The set of augmentations was optimized using a local search strategy. One augmentation policy at a time was introduced to the baseline SimCLR policies, and the performance was evaluated. All models were trained for 100 epochs total where one epoch consists of training on a random 30\% sub-sample --- roughly 3.8M unique samples --- of the full unlabeled dataset, this sample is redrawn every 20 epochs, allowing for reduced computational cost while still exposing the model to the majority of the full unlabeled data at some point during training. All models are trained using 4 V100-32GB GPUs, 16 CPU cores, and 200GB of RAM with a global batch size of 512, taking about 6 days to complete 100 epochs.  The resulting models are then evaluated on the classification task and the $H_{s}$ regression task. Figure~\ref{fig:auroc_samples_classification} shows the MLP transfer performance of different embeddings on the classification task for varying numbers of labeled training samples.
The results show that Mixup and CVAug consistently improve performance. However, the domain-inspired notch filter policy did not improve performance, and thus was not included in the final model. Detailed results are presented in Appendix~\ref{appendix_augmentations}.

\subsection{Transfer learning}

\begin{table}[ht]
    \begin{tabular}{@{}llcccccc@{}}
    \toprule
    \multicolumn{1}{c}{Model} & \multicolumn{1}{c}{Evaluation method} & \multicolumn{2}{c}{Classification}  & \multicolumn{2}{c}{Wave height (m)}     & \multicolumn{2}{c}{Air temperature (\textdegree C)} \\ \cmidrule(lr){3-4}\cmidrule(lr){5-6}\cmidrule(lr){7-8} 
                              &                                       & AUROC          & F1-Score       & MAE              & RMSE             & MAE              & RMSE             \\ \midrule
    \multicolumn{1}{c}{}      & ImageNet                              & 0.925          & 0.675          & - & - & - & - \\
    kNN                       & Baseline SimCLR                       & 0.925          & 0.669          & - & - & - & - \\
                              & WV-Net (ours)                          & \textbf{0.936} & \textbf{0.697} & - & - & - & - \\
                              &                                       &                &                &                  &                  &                  &                  \\
                              & ImageNet                              & 0.952          & 0.730          & 0.447            & 0.601            & 0.682            & 0.974            \\
    Linear                    & Baseline SimCLR                       & 0.954          & 0.739          & 0.395            & 0.532            & 0.655            & 0.920            \\
                              & WV-Net (ours)                          & \textbf{0.958} & \textbf{0.754} & \textbf{0.370}   & \textbf{0.500}   & \textbf{0.637}   & \textbf{0.902}   \\
                              &                                       &                &                &                  &                  &                  &                  \\
                              & ImageNet                              & 0.931          & 0.715          & 0.479            & 0.656            & 0.702            & 0.996            \\
    MLP                       & Baseline SimCLR                       & 0.930          & 0.716          & 0.355            & 0.491            & \textbf{0.691}   & \textbf{0.960}   \\
                              & WV-Net (ours)                          & \textbf{0.948} & \textbf{0.744} & \textbf{0.335}   & \textbf{0.459}   & 0.763            & 1.01             \\
                              &                                       &                &                &                  &                  &                  &                  \\
                              & ImageNet                              & 0.931          & 0.759          & 2.696            & 3.001            & 0.661            & 0.964            \\
    Finetuned                 & Baseline SimCLR                       & 0.934          & 0.760          & 0.418            & 0.586            & \textbf{0.623}   & \textbf{0.902}   \\
                              & WV-Net (ours)                          & \textbf{0.939} & \textbf{0.777} & \textbf{0.377}   & \textbf{0.530}   & 0.635            & 0.923           \\ \bottomrule
    \end{tabular}
    \caption{Comparison of final model performances. AUROC and F1 scores correspond to the image classification and MAE and RMSE are used for the regression tasks to estimate significant wave heights and air-sea temperature differences. The best score for each task under different evaluation scenarios is highlighted in bold.}  
    \label{tab:Tab02}  
\end{table}

Based on the optimization experiments above, we selected four augmentations to add to the baseline SimCLR augmentation pool: mixup, random color inversion, random rotation, and a random sharpness transform. The parameterization of these transforms remains unchanged. These are used to train the final model, called WV-Net. WV-Net is then compared to the baseline SimCLR model trained on WV images (without additional augmentations) and the ImageNet model trained using supervised learning. The two SSL models (WV-Net and baseline SimCLR) are pretrained for 200 epochs with a global batch size of 1024 (and accordingly a learning rate of 1.2) on 8 V100-32GB GPUs, using 400GB of RAM and 36 CPU cores. Training takes about 12 days to complete.

Table~\ref{tab:Tab02} compares the performance of the three models on three supervised learning tasks using four protocols. WV-Net outperforms the other models on most tasks under most evaluation scenarios. The only task where other models perform better than WV-Net is the air temperature prediction task, where WV-Net performs the worst in the MLP scenario and slightly worse than the baseline SimCLR model when finetuned end-to-end. In general, the linear models perform the best on the classification task while the MLP and finetuned models perform the best on the wave height and air temperature regression tasks respectively. 

Figures comparing WV-Net and ImageNet ROC curves and scatter plots for all tasks are provided in Appendix~\ref{appendix_performance}. The AUROC plots in Figure~\ref{fig:FigA2} illustrate that while both finetuned models exhibit excellent performance, WV-Net results in a slightly more robust model with fewer classes falling below a 0.9 AUROC.
The scatter plot for wave height regression in Figure~\ref{fig:FigA3} again shows the ImageNet weights failing to converge for this task. Notably, that is after limited hyperparameter tuning to have the majority of models fit the regression problems.
Lastly, Figure~\ref{fig:FigA4} shows the comparison for the air temperature regression task. Both models tend toward the mean but the WV-Net predictions are noticeably more well-distributed. We expect performance could be improved with more extensive task-specific hyperparameter tuning.

\subsection{Image retrieval}
The image-retrieval task illustrates the capability of the learned embeddings from the SSL model to delineate between features of interest without any finetuning. WV-Net outperforms ImageNet embeddings in almost all of the rare classes and remains competitive in all other cases, as detailed in Table~\ref{tab:Tab03}. Because the dataset is multilabel and several classes can be present in a single image, identifying all classes from a single example can be noisy and lead to mAP scores that appear lower than for single-label datasets common in natural image applications. Instead of scoring for any class overlap between the anchor and retrieved images, mAP for both models approaches 1.0, illustrating that they can retrieve images that share some dominant characteristics. The fact that WV-Net otherwise outperforms ImageNet suggests that the SSL embeddings are more sensitive to secondary classes present in the images, allowing for more fine-grained delineation. 

\begin{table}[ht]
    \begin{tabular}{@{}lccccccc@{}}
    \toprule
    Model        & \begin{tabular}[c]{@{}c@{}}AW \\ (N=101)\end{tabular} & \begin{tabular}[c]{@{}c@{}}IW \\ (N=304)\end{tabular} & \begin{tabular}[c]{@{}c@{}}OE \\ (N=142)\end{tabular} & \begin{tabular}[c]{@{}c@{}}SI \\ (N=955)\end{tabular} & \begin{tabular}[c]{@{}c@{}}IB \\ (N=762)\end{tabular} & \begin{tabular}[c]{@{}c@{}}SH \\ (N=236)\end{tabular} & \begin{tabular}[c]{@{}c@{}}SW \\ (N=167)\end{tabular} \\ \midrule
    ImageNet     & 0.013                                                 & 0.184                                                 & \textbf{0.130}                                        & 0.845                                                 & 0.223                                                 & 0.016                                                 & \textbf{0.024}                                        \\
    WV-Net (Ours) & \textbf{0.127}                                        & \textbf{0.297}                                        & 0.119                                                 & \textbf{0.901}                                        & \textbf{0.398}                                        & \textbf{0.021}                                        & 0.020                                                \\ \bottomrule
    \end{tabular}
    \caption{Comparison of image retrieval performance. mAP scores shown on rare classes in the GOALI dataset for ImageNet and WV-Net embeddings with the better-performing model for each class highlighted in bold.}  
    \label{tab:Tab03}  
\end{table}

\begin{figure}[ht]
  \centering
  \includegraphics[width=\textwidth]{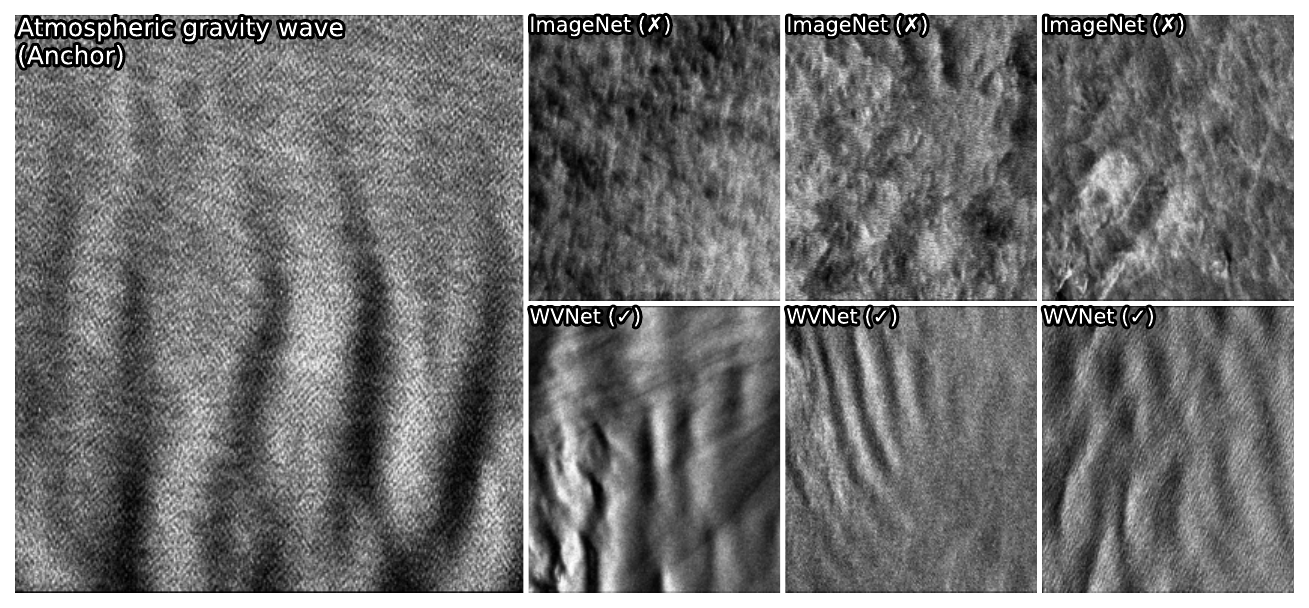}
  \caption{Image retrieval example for atmospheric gravity wave class. Anchor image (left column) is the query for kNN retrieval and the six images to the right are top-3 neighbors from ImageNet and WV-Net embeddings. This example shows successful image retrieval with the class present in the lower half of the anchor image.}
  \label{fig:Fig05}
\end{figure}

Figure~\ref{fig:Fig05} shows an example of retrieved images for a reference atmospheric gravity waves (AW) image, or anchor. This class makes up less than 0.5\% of the overall dataset. Given the anchor image on the left of Figure~\ref{fig:Fig05} from this class, WV-Net embeddings give an average precision of 0.95 for 20 retrieved images, outperforming the 0.11 average precision of ImageNet embeddings. It appears that the samples retrieved using ImageNet mostly share similar contrasts in the SAR backscatter, while WV-Net consistently identifies the correct characteristics associated with the class. However, Figure~\ref{fig:FigA5} in Appendix~\ref{appendix_performance} illustrates that given an anchor image where the class is less obvious (AB with subtle AW signatures). Nevertheless, Table~\ref{tab:Tab03} again shows that, on average, WV-Net is more robust to the anchor choice for the AW and most other classes. This is similar to the uncertainty that humans have when characterizing images that contain multiple features. This may also explain the overall low mAP scores shown for the SH (ship) and SW (ship wake) classes in Table~\ref{tab:Tab03}, because these are generally small, isolated objects in the image where other phenomena dominate the ocean surface backscatter.

\section{Limitations}

One major limitation of this work is computational constraints. Model performance could likely be improved with more larger models, longer training, and more extensive hyperparameter sweeps. Carefully tuning the temperature parameter during pretraining can impact task performance, especially for relatively small batch sizes such as those employed here~\cite{democratizingjessl_bordes_2023}. Contrastive SSL models have been shown to scale effectively with model capacity~ \cite{simclr_chen_2020, simclrv2_chen_2020}, thus we expect that training a larger model such as ResNet152(x2, x4) with our setup would result in even better performance on downstream tasks. 

Similarly, masked-image-modeling and variational inference with adversarial learning (ViTs) have been shown to outperform convolutional architectures given enough training time and data~\cite{mae_he_2021, mlpmixer_tolstikhin_2021}. While ViTs were included in the initial model analysis (Appendix~\ref{appendix_contrastive}), the models were relatively small. It is possible that given a larger ViT model and enough training time this could be a competitive approach to the one presented here. 

The downstream tasks presented are only a small subset of potential applications. For example, models could be trained to detect the organized large-scale eddies or lack of them (NV, WS, and MC) which are present in nearly 85\% of all images. Supplementing the results with a dense prediction task like detecting organized large-scale eddies could provide further insights into the behavior of WV-Net. Previous works such as~\citet{SikoraEtAl1995,YoungEtAl2002,YoungEtAl2005} have based their analysis of the physical dynamics associated with the marine atmospheric boundary layer (MABL) on hundreds of SAR images. WV-Net could help change the study of the MABL by systematically mapping millions of observations in time and space, changing the field from data-poor to data-rich. Even rarely occurring observations such as small-scale eddies ($<100$ m), atmospheric gravity waves in the open ocean, or lines in the sea~\cite{YoderEtAl1994} can be well-detected by WV-Net with minimal additional annotations. 




\section{Conclusion}
Using self-supervised contrastive learning on almost 10 million images, we have created  WV-Net, the first foundation model for S-1 WV imagery. Experiments on downstream classification, regression, and image-retrieval tasks support the two hypotheses: (1) a model pretrained with self-supervised contrastive learning on unannotated domain-specific imagery outperforms models pretrained with supervised learning on natural images, and (2) self-supervised contrastive models can further be improved for non-natural-image tasks by carefully selecting pretext tasks, or augmentations. However, we found that the best augmentation strategies were not necessarily the ones that leveraged any particular domain knowledge, such as random notch filtering. Instead, we found that the best augmentations were the original SimCLR augmentations plus mixup, rotations, color inversions, and sharpness transforms. 

While the performance improvement of WV-Net over ImageNet models is small for some tasks, the advantage is consistent across tasks. Of the three supervised learning tasks, the largest performance improvement is observed for the wave height prediction task, which requires extracting fine-scale features that are likely washed out in the ImageNet model. Furthermore, experiments demonstrate that WV-Net embeddings can yield state-of-the-art performance without the need for end-to-end finetuning, drastically reducing the need for computational resources and time. WV-Net is also more data-efficient than competing approaches, requiring less labeled data and even displaying strong image retrieval performance with no labeled data. Together, these properties make WV-Net a valuable tool for the remote sensing research community. WV-Net weights and code to run the model will be made available at \url{https://github.com/hawaii-ai/WVNet/}.

More generally, this work demonstrates the value of designing domain-specific foundation models. While WV-Net is designed specifically for WV-mode images from the Sentinel-1 mission, our approach can be applied to other remote sensing imaging technologies with different physical scales. These include other important ocean monitoring technologies like surface water and ocean topography (SWOT), other SAR modes, or scatterometers. Our experiments show the value of designing a pretext task that is appropriate for the domain, highlighting the value of close collaboration between machine learning and domain scientists.


\begin{ack}
RF and DV were supported by NASA Physical Oceanography grants NNX17AH17G and 80NSSC20K0822. JS was supported through grant number 2132150 from the National Science Foundation and NASA Physical Oceanography grant 80NSSC20K0822. YG was supported by the NASA Physical Oceanography grant 80NSSC20K0822. AM and BC were supported by ESA Contract No. 4000135827/21/NL - Harmony Science Data Utilisation and Impact Study for Ocean. AM was also supported by ESA Sentinel-1 Mission Performance Center 465 (4000107360/12/I-LG). We thank ESA for providing the data and IFREMER for computing resources used in this study.
LMW was supported by the National Science Foundation Graduate Research Fellowship Program under Grant No. 2236415.
Computing resources funded in part by NSF CC* awards \#2201428 and \#2232862 are gratefully acknowledged.
\end{ack}

{
\small
\bibliographystyle{plainnat}
\bibliography{main}
}


\appendix

\section{Comparison of contrastive learning frameworks and backend models}\label{appendix_contrastive}

Since there are multiple possible choices for contrastive self-supervised frameworks, we chose to evaluate one representative member of each of the framework families proposed by~\citet{sslcookbook_bakestriero_2023}. SimCLR~\cite{simclr_chen_2020} for the deep metric learning family, bootstrap your own latent (BYOL)~\cite{byol_grill_2020} for the self-distillation family, and swapping assignments between multiple views of the same image (SwAV)~\cite{swav_caron_2021}. 
Similarly, there are multiple potential choices for families of backend architecture that have shown promise in a broad range of computer vision tasks, we chose a ResNet50~\cite{resnet_he_2015} to represent a standard convolutional architecture, ConvNeXt-T~\cite{convnext_liu_2022} to represent a more modern version of a convolutional architecture, and a ViT-S/16~\cite{vit_dosovitskiy_2021} to represent vision transformers. The model sizes were chose to have roughly the same number of trainable parameters and constrained to fit the available compute budget. All hyperparameters were set in accordance with the original framework papers. 
Table~\ref{tab:TabA1} details the performance results for finetuned models on the classification and wave height tasks, showing clear dominance by the SimCLR + ResNet combination.

\begin{table}[!htb]
    \begin{tabular}{@{}llcc@{}}
    \toprule
    Framework & Backend Architecture & Classification (AUROC) & Wave height (RMSE) \\ \midrule
              & ResNet50             & \textbf{0.935}         & \textbf{0.564}     \\
    SimCLR    & ConvNeXt             & 0.911                  & 0.991              \\
              & ViT                  & 0.881                  & 1.153              \\
              &                      &                        &                    \\
              & ResNet50             & 0.931                  & 0.922              \\
    BYOL      & ConvNeXt             & 0.889                  & 1.133              \\
              & ViT                  & 0.885                  & 1.155              \\
              &                      &                        &                    \\
              & ResNet50             & 0.925                  & 0.780              \\
    SwAV      & ConvNeXt             & 0.915                  & 1.126              \\
              & ViT                  & 0.927                  & 1.168              \\ \bottomrule
    \end{tabular}    
    \caption{Validation set performance of different contrastive framework and backend architecture combination. Best model per task is highlighted in bold.}
    \label{tab:TabA1}
\end{table}

\section{SAR data processing}\label{appendix_sar}

Sentinel-1 launched two satellites, S-1 A and B, in April 2014 and 2016 respectively~\cite{sar_torres_2012}. A third, S-1 C, is scheduled to be launched in November 2024. S-1B went out of commission in December 2021. The S‐1 satellites are identical polar-orbiting, sun‐synchronous satellites~ \cite{TorresEtAl2012}. S-1 operates in the C‐band SAR with a center frequency of 5.405 GHz or wavelength of 5.5 cm. S-1 has a 12-day repeat cycle, flies with an altitude of 690 km, has an inclination of 98.2$^{\circ}$, and a repeat period of 98.7 minutes. When both S-1A and S-1B were in operation they were 180$^\circ$ out of phase equating to a 6-day repeat cycle. 

Each satellite produces approximately 60,000 images per month. S-1A and went into routine acquisition mode in October 2015 and July 2016 respectively, so in total, there are approximately 165 months and 9.9M S-1A/B images. The WV images are 20x20 km scenes and alternate between incidence angles of 23.8$^\circ$ (WV1) and 36.8$^\circ$ (WV2). The along-track separation is 100 km with 5 m pixel spacing. S-1 uses both vertical-vertical (VV) or horizontal-horizontal (HH) polarization, but only one polarization can be obtained for one image. The majority of the WV archive is in VV.

The S-1 \textit{geotiff} are saved in range-azimuth coordinates. The images in this study have the North direction facing upwards; therefore, the descending passes are flipped in the range and azimuth directions to make their relative geophysical representation the same. The raw 20-km WV images have 4000 to 5000 pixels in the range and azimuth directions. We implement a similar strategy to~ \cite{WangEtAl2019a} by reducing the raw data size while highlighting the geophysical phenomena that influence the sea surface roughness. The scales resolved by this processing are larger than the typical azimuth cutoff of 100 m~ \citet{StopaEtAl2015} and extend to 5 km in three steps: 1) incidence normalization, 2) downscaling, and 3) intensity normalization.
\begin{enumerate}
    \item \textit{Incidence Normalization:}
    The radar backscatter ($\sigma _{0}$) is strongly related to the local surface wind, incidence angle ($\phi$), relative wind-platform angle ($\theta$), and polarization. CMOD5N of~ \citet{Hersbach2010} is used to remove these effects by assuming a constant wind speed of 10 ms$^{-1}$ and a relative wind-platform angle $(\phi)$ of 45$^\circ$ to estimate the sea surface roughness (SSR) as
    \begin{equation}
     SSR=\frac{ \sigma _{0}}{CMOD5N\left(10~ms^{-1},\theta,\phi=45^{\circ},VV \right)}.
    \end{equation}
    
    \item \textit{Downscaling:}
    A moving boxcar window of 10x10 pixels or 50 m is applied to the $SSR$ data. Every 10th pixel is selected to reduce the data by a factor of 100 resulting in an image size of 400 to 500 pixels in both range and azimuth.
    
    \item  \textit{Intensity Normalization:}
    The image intensity is enhanced by normalizing each image with the 1st ($P01$) and 99th ($P99$) percentile 
    \begin{equation}
      SSR_{n}=255\left(\frac{ SSR-P01}{P99-P01}\right).
    \end{equation}
    This normalizes the $SSR$ to have values on the interval [0,255] where values of $SSR\leq P01=0$ and $SSR\geq =255$. Normalizing the values between [0,255] makes it effective to use an unsigned 8-bit integer and the matrix is saved as portable network graphics (\textit{png}). Note that the dataset is composed of grayscale images.
\end{enumerate}

\section{GOALI classes}\label{appendix_goali}

\begin{figure}[!htbp]
    \centering
    \includegraphics[width=\textwidth]{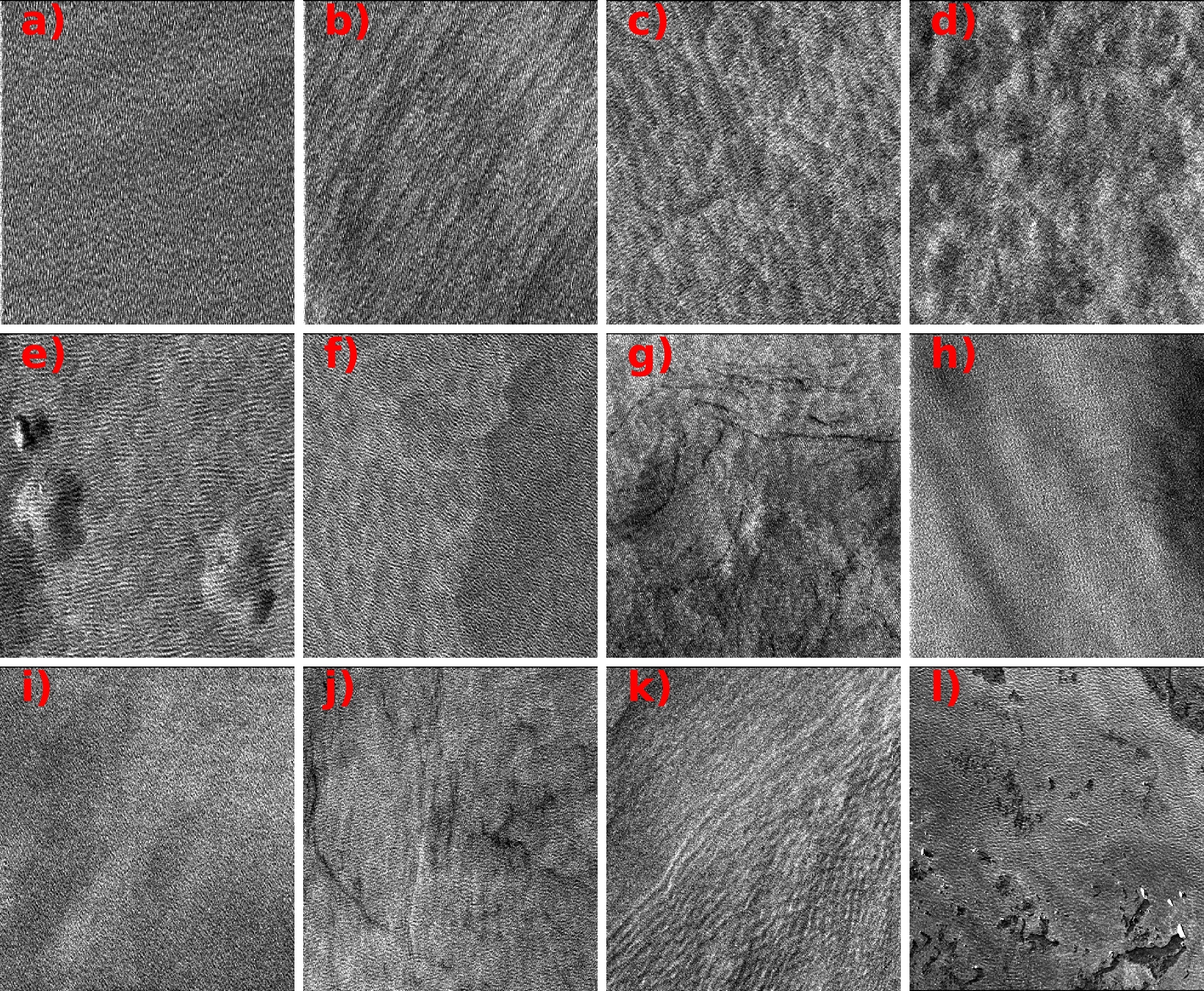}
    \caption[]{Twelve representative examples of the geophysical phenomena observed in the global Sentinel-1 WV archive. The panels are defined as negligible atmospheric variability (NV) (a), wind streaks (WS) (b), a mixture of WS and micro-scale cells (MC) (c), MC (d), rain cells (RC) also notice the two circular patterns in the bottom of the image that represent cold pools (CP) (e), sub-mesoscale air-mass boundary (AB) also contains WS/MC on the left-hand side (f), a low wind area (LW) containing biological slicks (the black lines) (BS) and MC, the circular structure of the BS are likely due to a small-scale eddy (g), atmospheric gravity waves (AW) (h), unidentified ocean or atmosphere (UD) (i), ocean front (OF) along with MC and BS (j), internal oceanic wave (k), and sea ice with icebergs (l).}
    \label{fig:sar_examples}
\end{figure}

Figure~\ref{fig:sar_examples} shows representative examples from each of the 12 GOALI classes. 

\section{Comparison of Augmentations}\label{appendix_augmentations}

This is a more detailed description of each introduced augmentation policy along with the reasoning behind why it may be beneficial to include in a WV-mode-specific contrastive learning model.
For the baseline SimCLR color jitter, since WV images are grey-scale, as part of image preprocessing the single greyscale channel is repeated three times and scaled within a 0-1. The images are then effectively treated as RGB throughout the augmentations and model. 

\paragraph{Cutout} is a geometric transform where one or multiple rectangles within the image are zeroed out or replaced with Gaussian noise~\cite{cutout_devries_2017} (see Figure~\ref{fig:sar_augmentations}i). We expect that including the cutout augmentation could, on one hand, replicate some of the driving factors behind the success of masked-image-modeling in geospatial data~\cite{mae_he_2021, satmae_cong_2022}, and also force the model to pay attention to all areas of the images despite the majority of the dataset being comprised of homogeneous textures across the 20-km frames. In this study, cutout can be applied up to three times, each application having a probability of $p=0.5$, each with a random scale between 2\% and 30\% of the image, and a random aspect ratio between 0.3 and 3.33. The areas may overlap and are zeroed out. 

\paragraph{CVAug} or computer vision augmentations is a composite of classical computer vision transforms that intuitively may complement learning on remote sensing data. Random color inversion is included to encourage a level of invariance to the pixel-intensity information and highlight textures. Random rotation is included because the model should be rotational invariant. A sharpness transform is included to obscure or highlight texture features, incentivizing more balanced representations that do not over-rely on global textures. Each augmentation is applied with probability $p=0.5$, the sharpness being increased by a factor of 0.5 and the rotation between $\pm 170^{\circ}$ (see Figure~\ref{fig:sar_augmentations}j).

\paragraph{Notch filtering} or stopband filters are common in signal processing, typically applied to reduce noise.~ \citet{spectrdropout_zhang_2017} use the term \textit{spectral dropout} to describe dropping weak Fourier coefficients from a layer's input distribution. Here, random dominant Fourier features from the raw input vector are dropped instead. Since ocean waves with scales of 50 to 800 m visually dominate most of the images, this augmentation should force the network to consider less dominant features. The notch filter is applied with probability $p=0.5$ and zeroes out up to 15 of the first 30 Fourier features obtained by doing a 2D Fourier transform over the image. However, the most dominant frequency, the first Fourier component, is excluded (see Figure~\ref{fig:sar_augmentations}k).

\paragraph{Mixup} was first proposed as a data augmentation for supervised learning~ \cite{mixup_zhang_2018}. It creates new training examples by taking a weighted combination of random feature vectors and their labels. Mixup has found popularity in SSL by only combining feature vectors such as in~ \cite{mixco_kim_2020, unmix_shen_2022, hardnegmixing_kalantidis_2020}. Further work framed mixup in terms of other noise-injection methods such as adding Gaussian noise to images, showing that it improves over random noise masks because the corrupted example is closer to the data manifold~\cite{dacl_verma_2021}. Mixup is applied with probability $p=0.5$ and a random mixup strength, $m$, between 0.1 and 0.4. Explicitly, the augmented image, $C$, created by mixup is written $C=(1-m)A+mB$ where $A$ is the original image and $B$ is another, randomly sampled, image from the same batch (see Figure~\ref{fig:sar_augmentations}l).

\paragraph{No-zoom crop} reduces the zooming component from the crop-and-zoom augmentation and is motivated by conserving the physical spatial scales within the satellite images. While the textures and objects can still vary in scale this is not as prominent as in natural images because the satellite is in a consistent orbit around the Earth and the WV images have a nearly fixed footprint of 20 km. Therefore, less pronounced zooming is intended to keep the characteristics of augmented images closer to what could be observed on the ocean's surface, ideally preventing the network from learning features on data enlarged so much that it has no bearing on downstream applications. In this setting, the cropped region can be no less than 90\% of the original image.

Table~\ref{tab:Tab01} gives the model performances trained with different augmentations for all transfer scenarios. Columns labeled classification show the performance on the multilabel image detection of oceanic and atmospheric phenomena. The columns labeled wave height show the performance for the significant wave height regression task. Note that the models labeled "Baseline" are the baseline SimCLR model with one added or removed augmentation. For classification, the baseline SimCLR models outperform the ImageNet transfer learning model for all evaluation criteria except the kNN and MLP F1 scores, where the ImageNet weights have a slight advantage. The ImageNet model performs well despite not being pretrained on SAR imagery. For regression, the finetuned ImageNet models did not converge to a competitive performance to predict wave heights. This could likely be overcome to an extent with careful, model-specific hyperparameter selection. This illustrates that transfer learning from models with larger domain gaps is sometimes more tuning-intensive than transferring weights from a model trained on similar data. Overall, this supports the hypothesis that even without domain-specific augmentations, simply training a self-supervised model on SAR WV data is an improvement over transferring weights from models trained on natural images. 

Further, the models highlighted in bold improve over the baseline SimCLR performance by adding augmentations. Across tasks and evaluation scenarios, the addition of various augmentations seems beneficial, but CVAug and mixup stand out for consistently outperforming both ImageNet weights and baseline SimCLR weights. While the classification scores of all models are relatively close (AUROCs $>$0.92 within 0.03), the wave height regression performance shows more pronounced differences with similar performance trends. Interestingly, while adding the cutout augmentation generally results in lower classification performance, it does benefit most regression models. The effects of the cutout augmentation are likely related to the scales of the phenomena - where ocean waves have typical scales of 100-600 m and many of the classified features are related to the atmosphere and have typical scales of 1-5 km.

Beyond absolute downstream performance, it is also important to understand how the models perform in data-constrained environments. An advantage of transfer learning is that it drastically reduces the need for labeled examples, lowering the barrier of entry for solving science problems with machine learning. Figure~\ref{fig:auroc_samples_classification} shows the MLP transfer performance of the differently pretrained models for low numbers of training samples for image classification. Mixup and CVAug robustly perform better than most other models with micro-AUROC statistics greater than 0.92 and only trained with 900 images. The performance differences are larger than those observed on the full training dataset in Table~\ref{tab:Tab01}. Therefore, for rare classes, or in situations when the training datasets are small, CVAug and mixup notably improve the model performance.

\begin{table}[ht!]
    \centering   
    \begin{tabular}{@{}llcccc@{}}
    \toprule
    \multicolumn{1}{c}{Model} & \multicolumn{1}{c}{Evaluation method} & \multicolumn{2}{c}{Classification}   & \multicolumn{2}{c}{Wave height} \\ \cmidrule(lr){3-4} \cmidrule(lr){5-6} 
                              &                                       & AUROC          & F1-Score       & MAE            & RMSE           \\ \midrule
    \multicolumn{1}{c}{}      & ImageNet                              & 0.925          & \textbf{0.675} & -              & -              \\
                              & Baseline SimCLR                       & 0.928          & 0.674          & -              & -              \\
                              & Baseline + Cutout                     & 0.920          & 0.660          & -              & -              \\
    kNN                       & Baseline + CVAug                      & \textbf{0.935} & \textbf{0.690} & -              & -              \\
                              & Baseline + Mixup                      & \textbf{0.930} & \textbf{0.676} & -              & -              \\
                              & Baseline + Notch                      & 0.926          & 0.668          & -              & -              \\
                              & Baseline - Zoom                       & 0.927          & 0.661          & -              & -              \\
                              &                                       &                &                &                &                \\
    \multicolumn{1}{c}{}      & ImageNet                              & 0.952          & 0.730          & 0.447          & 0.601          \\
                              & Baseline SimCLR                       & 0.953          & 0.735          & 0.428          & 0.580          \\
                              & Baseline + Cutout                     & 0.943          & 0.705          & \textbf{0.380} & \textbf{0.516} \\
    Linear                    & Baseline + CVAug                      & \textbf{0.954} & \textbf{0.742} & \textbf{0.413} & \textbf{0.555} \\
                              & Baseline + Mixup                      & \textbf{0.953} & \textbf{0.738} & \textbf{0.392} & \textbf{0.531} \\
                              & Baseline + Notch                      & 0.949          & 0.720          & \textbf{0.427} & \textbf{0.578} \\
                              & Baseline - Zoom                       & \textbf{0.955} & 0.722          & 0.485          & 0.647          \\
                              &                                       &                &                &                &                \\
    \multicolumn{1}{c}{}      & ImageNet                              & 0.931          & \textbf{0.715} & 0.479          & 0.656          \\
                              & Baseline SimCLR                       & 0.933          & 0.709          & 0.406          & 0.561          \\
                              & Baseline + Cutout                     & 0.919          & 0.681          & \textbf{0.360} & \textbf{0.494} \\
    MLP                       & Baseline + CVAug                      & \textbf{0.941} & \textbf{0.728} & \textbf{0.385} & \textbf{0.533} \\
                              & Baseline + Mixup                      & \textbf{0.938} & \textbf{0.725} & \textbf{0.373} & \textbf{0.514} \\
                              & Baseline + Notch                      & 0.924          & 0.694          & 0.410          & 0.564          \\
                              & Baseline - Zoom                       & 0.926          & 0.687          & 0.488          & 0.645          \\
                              &                                       &                &                &                &                \\
    \multicolumn{1}{c}{}      & ImageNet                              & 0.931          & 0.759          & 2.696          & 3.001          \\
                              & Baseline SimCLR                       & 0.932          & 0.765          & 0.424          & 0.604          \\
                              & Baseline + Cutout                     & \textbf{0.933} & 0.749          & 0.448          & 0.629          \\
    Finetuned                 & Baseline + CVAug                      & \textbf{0.935} & \textbf{0.768} & 0.433          & 0.614          \\
                              & Baseline + Mixup                      & \textbf{0.935} & \textbf{0.770} & \textbf{0.393} & \textbf{0.556} \\
                              & Baseline + Notch                      & \textbf{0.934} & 0.748          & \textbf{0.419} & \textbf{0.598} \\
                              & Baseline - Zoom                       & 0.931          & \textbf{0.765} & 0.444          & 0.624          \\ \bottomrule
    \end{tabular}
    \caption{Augmentation study results - showing the transfer performance of differently pretrained models. Note that rows 2 through 6 for each model setup correspond to WV-Net with different augmentations included. The classification scores are micro-averaged AUROC and micro-averaged F1-scores (higher is better).  The wave height scores are the RMSE and MAE (lower is better). Models that outperform baseline SimCLR are in bold.}  
    \label{tab:Tab01}  
\end{table}

\section{Evaluation protocols}\label{appendix_eval}

\subsection{Multilabel classification}

For all classification tasks a subset of classes from the GOALI dataset (WS, MC, NV, RC, CP, AB, LW, BS, OF, IW, and SI) will be considered as they comprise the majority the phenomena of interest for downstream applications and together comprise the vast majority of the dataset. All other labels are grouped into a catch-all "Other" class. The classification data is stratified and randomly split into 60\%, 20\%, and 20\% of the original 10,000 images for training, validation and hyperparameter tuning, and final model testing, respectively. All 6,400 images from~\citet{WangEtAl2019a} are held out for testing.

The kNN classifier is trained according to the protocol from~\cite{instancediscssl_wu_2018} with 15 neighbors, chosen based on a hyperparameter sweep. Cosine similarity (Equation~\ref{cos_sim_eq}) is used as the distance metric for the kNN model. 

The protocol from~\cite{simclr_chen_2020} is directly adopted for linear probing with no modifications.

The proposed MLP architecture from~\citet{democratizingjessl_bordes_2023} (2 hidden layers with 2048 ReLU~ \cite{relu_agarap_2019} units) is adopted and the model is trained for 200 epochs using the Adam 
optimizer~\cite{adam_kingma_2017} with a constant learning rate of 0.001. 

The fine-tuning procedure from~\cite{simclr_chen_2020} is followed with the batch size reduced to 256 and the learning rate scaled to 0.05 accordingly. 

The fine-tuned classification models are trained to minimize the sum of binary cross-entropy losses over all individual classes to allow for the multilabel property:
\begin{equation}
    \mathcal{L}_{cls} = -\sum_{i=1}^{N}(\sum_{c=1}^{C}y_{i,c}\log(\hat{y}_{i,c}) + (1 - y_{i,c})\log(1 - \hat{y}_{i,c})))
\end{equation}

The micro-averaged AUROC is computed by summing the predictions for each class and then calculating an AUROC curve for the aggregated predictions. The F1-score is related to the $precision$, or how many positive predictions made by the model were correct, and $recall$, or how many of the positive class samples present in the dataset were correctly identified by the model and calculated in terms of the true positives ($TP$), false positives ($FP$), an false negatives ($FN$) as
\begin{equation}
    F1=\frac{2}{\frac{1}{precision}+\frac{1}{recall}}=\frac{TP}{TP+0.5(FP+FN)}.
\end{equation}

\subsection{Image retrieval}

For the image retrieval task, 100 experiments are conducted on IW, OE, SI, AW, IB, SH, and SW, the rarest classes from the combined classification dataset. For each experiment a random sample of each class is drawn and then the 20 closest neighbors are retrieved based on the embeddings generated by an ImageNet model and our WV-Net model with no supervised training on the labeled data. mAP is calculated for each class and then averaged over all classes. 

\subsection{Regression}

Linear probing and MLP protocols are left unchanged from the classification protocols except for adjusting the output function. 
Hyperparameters were minimally adjusted for the end-to-end finetuning scenario since the hyperparameters used for the classification task showed instability. The final parameters are $10^{-6}$ weight decay, a backbone learning rate of 0.007, an output-layer learning rate of 0.025, and a dropout rate of 0.5. All other hyperparameters are left unchanged.
The finetuned regression models use a softplus output unit and are trained using a weighted combination of the mean absolute (or L1) error (MAE) and the mean squared error (MSE) weighted in favor of the MSE: 
\begin{equation}
    \mathcal{L}_{reg} = \sum_{i=1}^{N}0.1*|y_i-\hat{y}_i| + (y_i-\hat{y}_i)^2
\end{equation}

All regression models are primarily evaluated using the root mean squared error (RMSE) and MAE.

\section{WV-Net performance details}\label{appendix_performance}

\begin{figure}[!htbp]
    \centering
    \includegraphics[width=\textwidth]{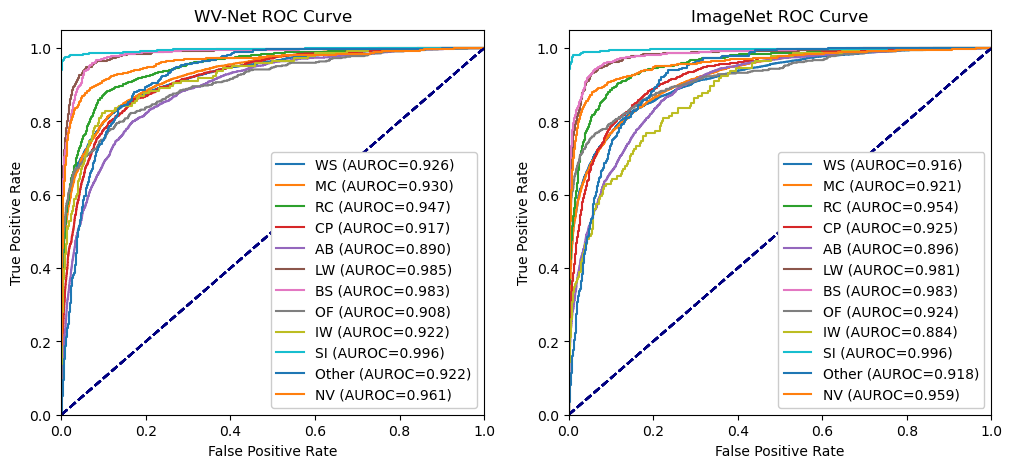}
    \caption{ROC curve comparison for WV-Net and ImageNet models finetuned on multilabel classification task.}
    \label{fig:FigA2}
\end{figure}

\begin{figure}[!htbp]
    \centering
    \includegraphics[width=\textwidth]{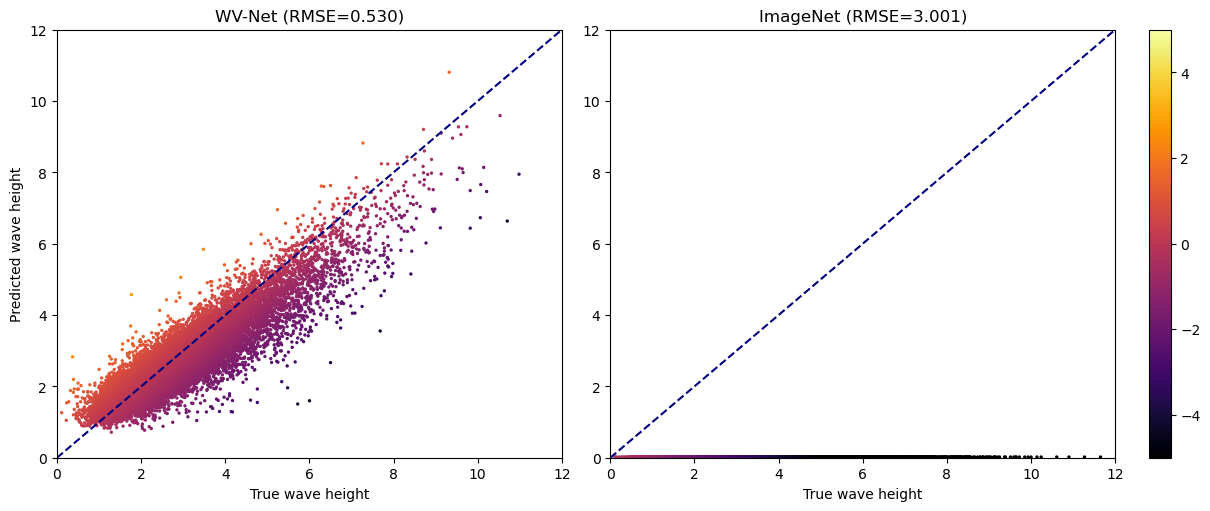}
    \caption{Scatter plot comparison for WV-Net and ImageNet models finetuned for the wave height regression task. Note that the ImageNet weights failed to converge even with limited hyperparameter tuning for this task.}
    \label{fig:FigA3}
\end{figure}

\begin{figure}[!htbp]
    \centering
    \includegraphics[width=\textwidth]{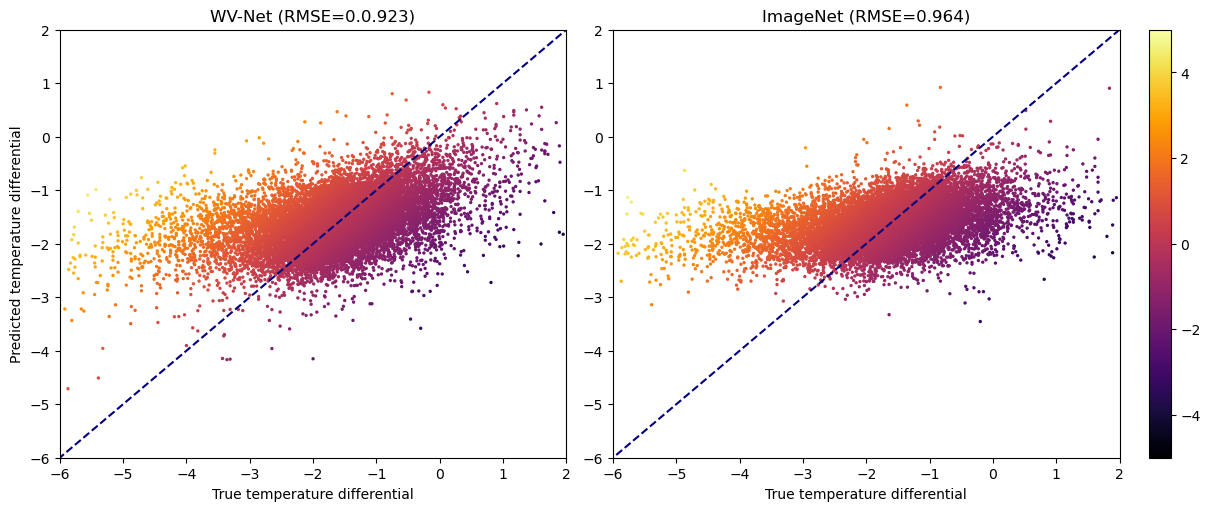}
    \caption{Scatter plot comparison for WV-Net and ImageNet models finetuned for the air temperature regression task.}
    \label{fig:FigA4}
\end{figure}

\begin{figure}[ht]
  \centering
  \includegraphics[width=\textwidth]{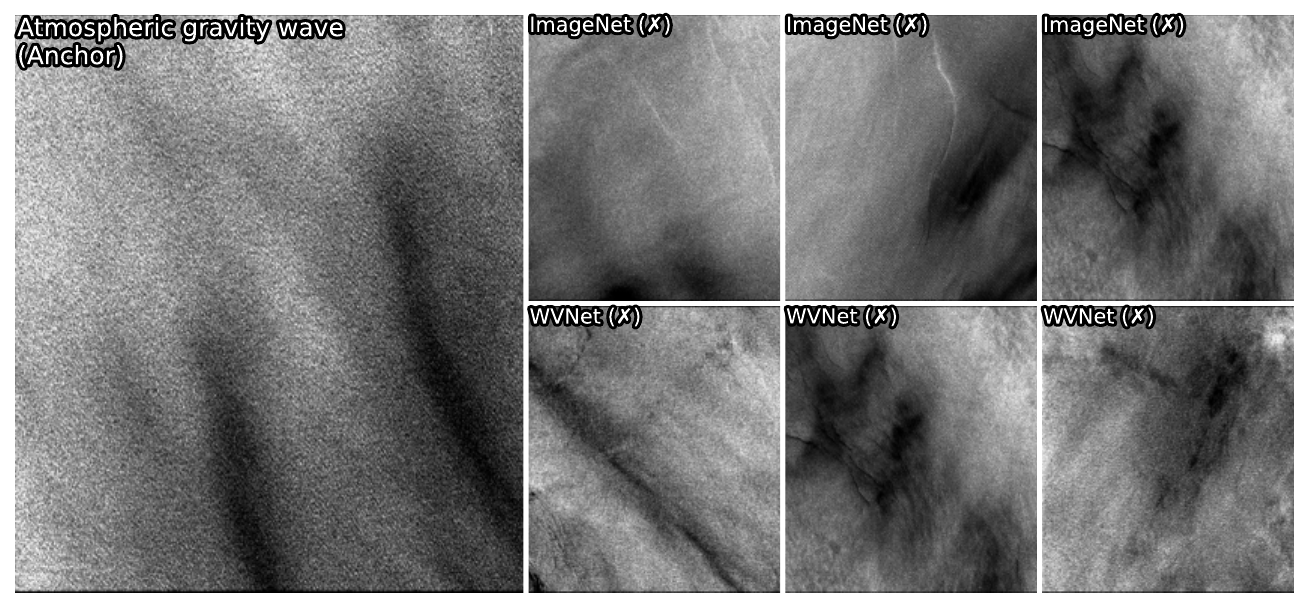}
  \caption{Image retrieval example for atmospheric gravity wave showing unsuccessful image retrieval with the class hard to discern in the anchor image. This sample illustrates an anchor for which both architectures have uncertainty since the target class in the anchor image is not well pronounced.}
  \label{fig:FigA5}
\end{figure}

\end{document}